\documentclass[runningheads]{llncs}
\usepackage{graphicx}

\newcommand{\citep}{\cite}
\newcommand{\citet}{\cite}
\newcommand\bycite[1]{in~\citet{#1}}

\newif\ifwithappendix
\withappendixfalse

\newif\ifappendixshown
\appendixshowntrue

\usepackage[T1]{fontenc}
\usepackage[utf8]{inputenc}
\usepackage[textsize=tiny]{todonotes}
\usepackage{graphicx}
\usepackage{booktabs}
\usepackage{latexsym}

\usepackage{footnote}
\makesavenoteenv{tabular}
\makesavenoteenv{table}
\makesavenoteenv{table*}

\newcommand\minput[1]{%
  \input{#1}%
  \ifhmode\ifnum\lastnodetype=11 \unskip\fi\fi}

\usepackage{hyperref}
\usepackage{xstring}

\newcommand{\code}[1]{\texttt{#1}}
\newcommand{\noqa}[1]{}
\newcommand{\noqall}[1]{}

\usepackage{amsmath}

\usepackage{multirow}
\usepackage{longtable}

\newcommand{\entity}[1]{{\small{\texttt{#1}}}}

\usepackage{pifont}
\newcommand{\cmark}{\textcolor{green}{\ding{51}}}
\newcommand{\xmark}{\textcolor{red}{\ding{55}}}

\begin{document}

\title{Kleister: Key Information Extraction Datasets \\Involving Long Documents with Complex Layouts}

\author{
    Tomasz Stanisławek\inst{1,2}  \and
    Filip Graliński\inst{1,3}  \and
    Anna Wróblewska\inst{2}  \and\\
    Dawid Lipiński\inst{1}  \and
    Agnieszka Kaliska\inst{1,3}  \and
    Paulina Rosalska\inst{1}  \and\\
    Bartosz Topolski\inst{1}  \and
    Przemysław Biecek\inst{2,4}
    }

\authorrunning{T. Stanisławek et al.}

\institute{
    Applica.ai, 15 Zajęcza, Warsaw, 00351 \email{firstname.lastname@applica.ai}  \and
    Warsaw University of Technology, Koszykowa 75, Warsaw, Poland \email{firstname.lastname@pw.edu.pl}  \and
    Adam Mickiewicz University, 1 Wieniawskiego, Poznan, 61712, Poland \email{firstname.lastname@amu.edu.pl}  \and
    Samsung R\&D Institute Poland, Plac Europejski 1, Warsaw, Poland \email{firstnameletter.lastname@samsung.com}
    }

\maketitle              %
\begin{abstract}
The relevance of the Key Information Extraction (KIE) task is increasingly important in natural language processing problems. But there are still only a few well-defined problems that serve as benchmarks for solutions in this area. To bridge this gap, we introduce two new datasets (\emph{Kleister NDA} and \emph{Kleister Charity}). They involve a mix of scanned and born-digital long formal English-language documents. In these datasets, an NLP system is expected to find or infer various types of entities by employing both textual and structural layout features. The Kleister Charity dataset consists of 2,788 annual financial reports of charity organizations, with 61,643 unique pages and 21,612 entities to extract. The Kleister NDA dataset has 540 Non-disclosure Agreements, with 3,229 unique pages and 2,160 entities to extract. We provide several state-of-the-art baseline systems from the KIE domain (Flair, BERT, RoBERTa, LayoutLM, LAMBERT), which show that our datasets pose a strong challenge to existing models. The best model achieved an 81.77~\% and an 83.57~\% F1-score on respectively the Kleister NDA and the Kleister Charity datasets. We share the datasets to encourage progress on more in-depth and complex information extraction tasks.

\end{abstract}

\noqall{proselint-consistency.spacing}

\noqall{spell-pdf2djvu}
\noqall{spell-djvu2hocr}
\noqall{spell-Kleister}
\noqall{spell-SROIE}
\noqall{spell-LayoutLM}
\noqall{spell-LAMBERT}
\noqall{spell-KIE}
\noqall{spell-autotagging}
\noqall{spell-autotagged}

\section{Introduction}

The task of Key Information Extraction (KIE) from Visually Rich Documents (VRD) has proved increasingly interesting in the business market with the recent rise of solutions related to Robotic Process Automation (RPA). From a business user's point of view, systems that, fully automatically, gather information about individuals, their roles, significant dates, addresses and amounts, would be beneficial, whether the information is from invoices or receipts, from company reports or contracts~\cite{Wellmann_2020,park2019cord,Liu_2019,Palm_2019,holt-chisholm-2018-extracting,DBLP-journals-corr-abs-1809-08799,DBLP-conf-fedcsis-WroblewskaSPG18}.
There is a disparity between what can be delivered with the KIE domain systems on publicly available datasets and what is required by real-world business use. This disparity is still large and makes a robust evaluation difficult.
Recently, researchers have started to fill the gap by creating datasets in the KIE domain such as scanned receipts: \emph{SROIE}\footnote{\url{https://rrc.cvc.uab.es/?ch=13\&com=evaluation\&task=3}} \cite{park2019cord}, form understanding~\cite{jaume2019}, NIST Structured Forms Reference Set of Binary Images (\emph{SFRS})\footnote{\url{https://www.nist.gov/srd/nist-special-database-2}} or Visual Question Answering dataset \emph{DocVQA}~\cite{mathew2021docvqa}.

\begin{figure}[t]
\includegraphics[width=\columnwidth]{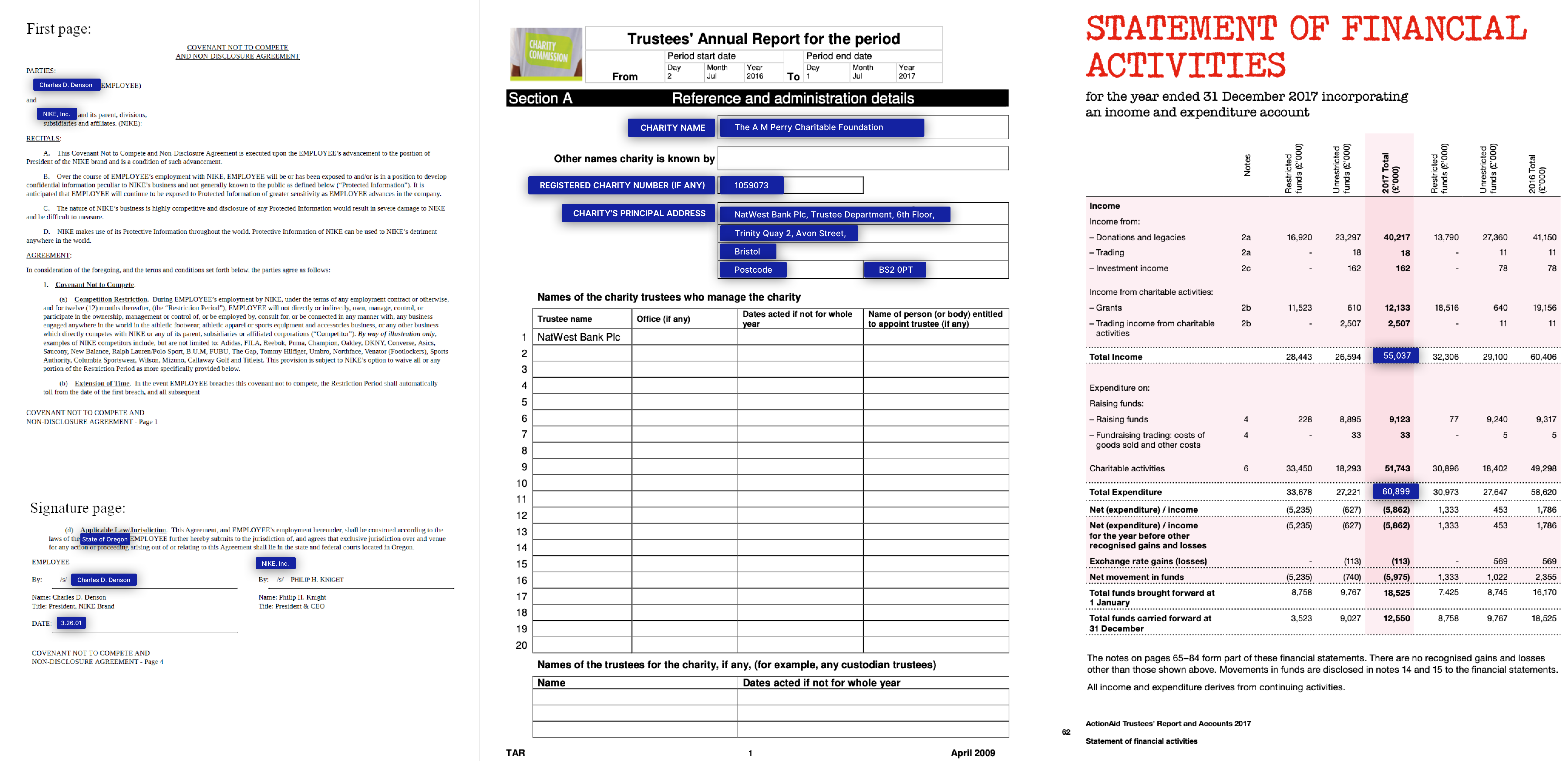}
\caption{Examples of a real business applications and data for \emph{Kleister} datasets. (Note: The key entities are in blue.)\label{fig:datasets_problems}}
\end{figure}

This paper describes two new English-language datasets for the Key Information Extraction tasks from a diverse set of texts, long scanned and born-digital documents with complex layouts, that address real-life business problems (Figure~\ref{fig:datasets_problems}). The datasets represent various problems arising from the specificity of business documents and associated business conditions, e.g. complex layouts, specific business logic, OCR quality, long documents with multiple pages, noisy training datasets, and normalization. Moreover, we evaluate several systems from the KIE domain on our datasets and analyze KIE tasks' challenges in the business domain. We believe that our datasets will prove a good benchmark for more complex Information Extraction systems.

The main contributions of this study are:
\begin{enumerate}
    \item \emph{Kleister} – two novel datasets of long documents with complex layouts: 3,328 documents containing 64,872 pages with 23,772 entities to extract (see Section~\ref{sec:dataset});
    \item our method of collecting datasets using a semi-supervised methodology, which reduces the amount of manual work in gathering data; this method has the potential to be reused for similar tasks (see Section~\ref{sec:NDA_dataset} and~\ref{sec:charity_dataset});
    \item evaluation over several state-of-the-art Named Entity Recognition (NER) architectures (Flair, BERT, RoBERTa, LayoutLM, LAMBERT) employing our \emph{Pipeline} method (see Section~\ref{sec:pipeline} and~\ref{sec:baseline_results});
   \item detailed analysis of the data and baseline results related to the Key Information Extraction task carried out by human experts (see Section~\ref{sec:statistics_analysis} and~\ref{sec:baseline_results}).
\end{enumerate}

The data
are available
at \url{https://github.com/applicaai/kleister-nda.git}
and \url{https://github.com/applicaai/kleister-charity.git}.

\section{Related Work}
\label{sec:review}

Our main reason for preparing a new dataset was to develop a strategy to deal with challenges faced by businesses, which means overcoming such difficulties as complex layout, specific business logic (the way that content is formulated, e.g. tables, lists, titles), OCR quality, document-level extraction and normalization.

\subsection{KIE from Visually Rich Documents (publicly available)}

A list of KIE-oriented challenges is available at the International Conference on Document Analysis and Recognition ICDAR 2019\footnote{\url{http://icdar2019.org/competitions-2/}} (cf. Table~\ref{tab:dataset_summary}). There is a dataset called SROIE\footnote{\url{https://rrc.cvc.uab.es/?ch=13}} with information extraction from a set of scanned receipts. The authors prepared 1,000 whole scanned receipt images with annotated entities: company name, date, address, and total amount paid (a similar dataset was also created \bycite{park2019cord}). Form Understanding in Noisy Scanned Documents is another interesting dataset from ICDAR 2019 \noqa{spell-FUNSD}(\emph{FUNSD})~\cite{jaume2019}. FUNSD aims at extracting and structuring the textual content of forms. However, the authors focus mainly on understanding tables and a limited range of document layouts, rather than on extracting particular entities from the data. The point is, therefore, to indicate a table but not to extract the information it contains.

\subsection{KIE from Visually Rich Documents (publicly unavailable)}

There are also datasets for the Key Information Extraction task based on invoices~\cite{Palm_2019,Palm_2017,holt-chisholm-2018-extracting,DBLP-journals-corr-abs-1809-08799}. Documents of this kind contain entities like ‘Invoice date,’ ‘Invoice number,’ ‘Net amount’ and ‘Vendor Name’, extracted using a combination of NLP and Computer Vision techniques. The reason for such a complicated multi-domain process is that spatial information is essential for properly understanding these kinds of documents. However, since they are usually short, the same information is relatively rarely repeated, and therefore there is no need for understanding the more extended context of the document. Nevertheless, those kinds of datasets are the most similar to our use case.

\subsection{Information Extraction from one-dimensional documents}

The \emph{WikiReading} dataset~\cite{hewlett-etal-2016-wikireading} (and its variant \emph{WikiReading Recycled}~\cite{dwojak-etal-2020-dataset}) is a large-scale natural language understanding task. Here, the main goal is to predict textual values from the structured knowledge base, Wikidata, by reading the text of the corresponding Wikipedia articles. Some entities can be extracted from the given text directly, but some have to be inferred. Thus, as in our assumptions, the task contains a rich variety of challenging extraction sub-tasks and it is also well-suited for end-to-end models that must cope with longer documents.

Key Information Extraction is different from the Named Entity
Recognition task (the \emph{CoNLL 2003} NER challenge~\cite{W03-0419} being a well-known
example). This is because: (1) retrieving spans is not required in KIE; (2) a system
is expected to extract specific, actionable data points rather than
general types of entities (such as people, organization, locations and
``others'' for CoNLL 2003).

\begin{table*}[ht!]
\centering
{\footnotesize{
\begin{tabular}{p{2.7cm}p{1.2cm}p{1.9cm}p{1.3cm}p{1.2cm}p{1.3cm}p{1.8cm}}
\toprule
Dataset name & CoNLL 2003 & WikiReading & \noqa{spell-FUNSD}FUNSD  & SROIE & Kleister NDA & Kleister Charity\\
\midrule
Source & Reuters news & Wikipedia & forms & receipts & EDGAR & UK Charity Com. \\
Documents & 1,393 & 4.7M\noqa{spell-7M} & 199 & 973 & 540 & 2,778  \\
Pages & — & — & 199 & 973 & 3,229 & 61,643  \\
Entities & 35,089 & 18M\noqa{spell-18M} & 9,743 & 3,892 & 2,160 & 21,612\\
\midrule
train docs & 946 & 16.03M & 149 & 626 & 254 & 1,729\\
dev docs & 216 & 1.89M & — & — & 83 & 440\\
test docs & 231 & 0.95M & 50 & 347 & 203 & 609\\
\midrule
Input/Output on token level(*) & \cmark & \xmark & \cmark & \cmark & \xmark & \xmark \\
\midrule
Long Document(*) & \xmark & \cmark & \xmark & \xmark & \cmark & \cmark \\
\midrule
Complex layout(*) & \xmark & \xmark & \cmark & \cmark & \cmark & \cmark \\
\midrule
OCR(*) & \xmark & \xmark & \cmark & \cmark & \xmark & \cmark \\
\bottomrule
\end{tabular}
}}
\vspace{1mm}
\caption{Summary of the existing English datasets and the Kleister sets. (*) For detailed description see Section~\ref{sec:statistics_analysis}.\label{tab:dataset_summary}}
\end{table*}

\section{Kleister: New Datasets}
\label{sec:dataset}

We collected datasets of long formal born-digital documents, namely US non-disclosure agreements (Kleister NDA) and a mixture of born-digital and (mostly) scanned annual financial reports of charitable foundations from the UK (Kleister Charity). These two datasets have been gathered in different ways due to their repository structures. Also, they were published on the Internet for different reasons. The crucial difference between them is that the NDA dataset was born-digital, but that the Charity dataset needed to be OCRed. Kleister datasets have a multi-modal input (PDF files and text versions of the documents) and a list of entities to be found.

\subsection{NDA Dataset}
\label{sec:NDA_dataset}

The NDA Dataset contains Non-disclosure Agreements, also known as Confidentiality Agreements. They are legally binding contracts between two or more parties, where the parties agree not to disclose information covered by the agreement. The NDAs can take on various forms (e.g. contract attachments, emails), but they usually have a similar structure.

\textbf{Data Collection Method.} The NDAs were collected from the Electronic Data Gathering, Analysis and Retrieval system (EDGAR\footnote{\url{https://www.sec.gov/edgar.shtml}}) via Google's search engine. The original files were in an HTML format, but they were transformed into PDF files to keep processing simple and similar to that of other public datasets. Transformation was made using the \texttt{puppeteer} library.\footnote{\url{https://github.com/puppeteer/puppeteer}} Then, a list of entities was established (see Table~\ref{tab:dataset_summary}).

\textbf{Annotation Procedure.} We annotated the whole dataset in two ways. Its first part, made up of 315~documents, was annotated by three annotators, except that only contexts with some similarity, pre-selected using methods based on semantic similarity (cf.~\citep{DBLP:conf/emnlp/BorchmannWGKJSP20}), were taken into account; this was to make the annotation faster and less-labor intensive. The second part, with 195 documents, was annotated entirely by hand. When preparing the dataset, we wanted to determine whether semantic similarity methods could be applied to limit the time it would take to perform annotation procedures; this solution was about 50~\% quicker than fully manual annotation. The annotations on all documents were then checked by a super-annotator, which ensured the annotation's excellent quality Cohen's $\kappa$ (=0.971)\footnote{\url{https://en.wikipedia.org/wiki/Cohen\%27s\_kappa}}.
Next, all entities were normalized according to the standards adopted
by us, e.g. the \entity{effective date} was standardized according to
ISO~8601 \noqa{spell-YYYY}i.e.~YYYY-MM-DD\footnote{The normalization standards are described in the public repository with datasets.}.

\textbf{Dataset split.} The Kleister NDA dataset contains a relatively small document count, so we decided to add more examples into the test split (about 38~\%) so as to be more accurate during the evaluation stage (see Table~\ref{tab:dataset_summary} for exact numbers).

\subsection{Charity Dataset}
\label{sec:charity_dataset}

The Charity dataset consists of annual financial reports that all charities registered in England and Wales must submit to the Charity Commission. The Commission subsequently makes them publicly available on its website.\footnote{\url{https://apps.charitycommission.gov.uk/showcharity/registerofcharities/RegisterHomePage.aspx}} There are no strict rules about the format of these charity reports. Some are richly illustrated with photos and charts and financial information constitutes only a small part of the entire report. In contrast, others are a few pages long and only necessary data on revenues and expenses in a given calendar year are given.

\begin{figure*}[h!t]
\begin{tabular}{cc}
 \includegraphics[width=5.3cm]{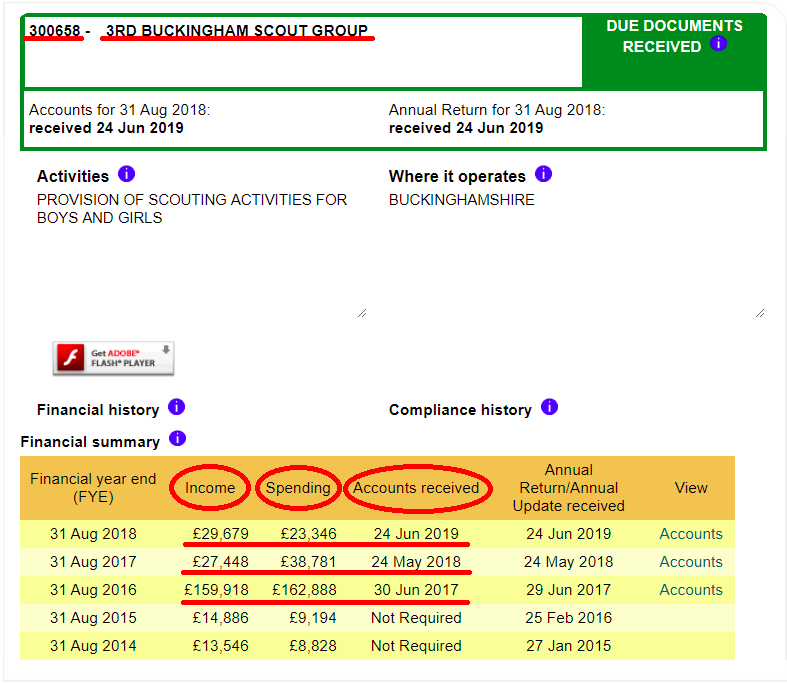}    &
 \includegraphics[width=5.3cm]{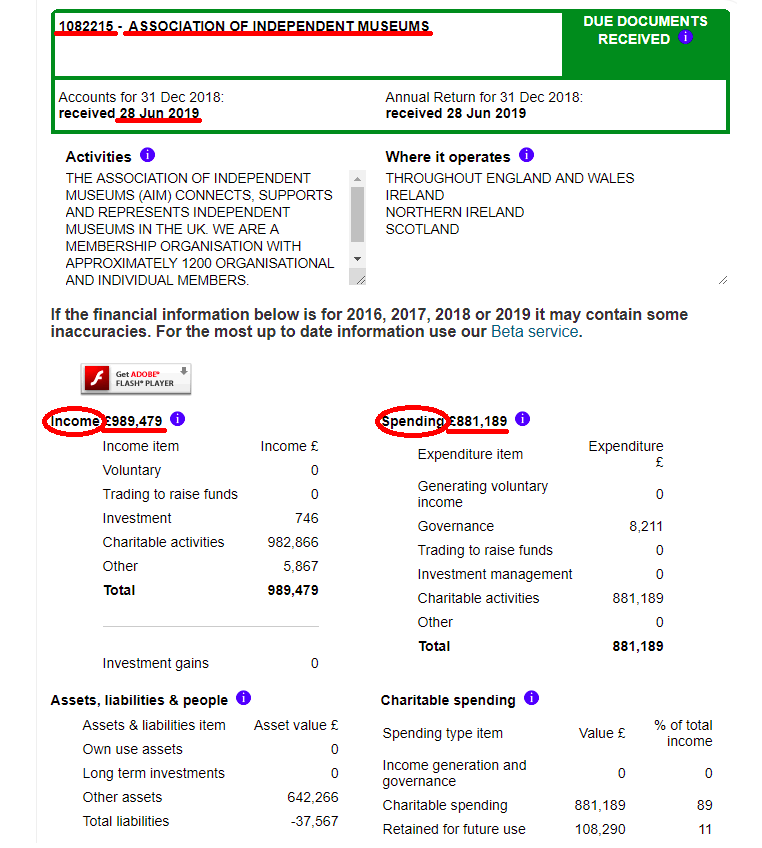}
\end{tabular}
\caption{Organization's page on the Charity Commission's website (left: organization whose annual income is between 25k\noqa{spell-25k} and 500k\noqa{spell-500k} GBP, right: over 500k). Note: Entities are underlined in red and names of entities are circled.\label{fig:charity-web-page}}
\end{figure*}

\textbf{Data Collection Method.} The Charity Commission website has a database of all the charity organizations registered in England and Wales. Each of these organizations has a separate sub-page on the Commission's website, and it is easy to find the most important information about them there (see Figure~\ref{fig:charity-web-page}). This information only partly overlaps with information in the reports. Some entities such as, say, a list of trustees might not be in the reports. Thus, we decided to extract only those entities which also appear in the form of a brief description on the website.

In the beginning, we downloaded 3,414 reports (as PDF files).\footnote{Organizations with an income below 25,000 GBP a year are required to submit a condoned financial report instead.} During document analysis, it emerged that several reports were written in Welsh. As we are interested only in English, all documents in other languages were identified and removed from the collection. Additionally, documents that contained reports for more than one organization or whose OCR quality was low were deleted. This left us with 2,778 documents.

\textbf{Annotation Procedure.} There was no need to manually annotate all documents because information about the reporting organizations could be obtained directly from the Charity Commission. Initially, only a random sample of 100 documents were manually checked. Some proved low quality: \entity{charity name} (5~\% of errors and 13~\% of minor differences), and \entity{charity address} (9~\% of errors and 63~\% of minor differences). Minor errors are caused by data presentation differences on the page and in the document. For example, the charity's name on the website and in the document could be written with the term \textit{Limited} (shortened to \textit{LTD}) or without it. These minor differences were corrected manually or automatically. In the next step, 366~documents were analyzed manually. Some parts of the charity's address were also problematic. For instance, counties, districts, towns and cities were specified on the website, but not in the documents, or \textit{vice versa}. We split the address data into three separate entities that we considered the most essential: postal code, postal town name and street or road name. The postal code was the critical element of the address, based on the city name and street name\footnote{Postal codes in the UK were aggregated from \url{www.streetlist.co.uk}}. The whole process allowed us to accurately identify entities (see Table~\ref{tab:dataset_summary}) and to obtain a good-quality dataset with annotations corresponding to the gold standard.

\textbf{Dataset split.} In the Kleister Charity dataset, we have multiple documents from the same charity organization but from different years. Therefore, we decided to split documents based on charity organization into the train/dev/test sets with, respectively, a 65/15/20 dataset ratio (see Table~\ref{tab:dataset_summary} for exact numbers). The documents from the dev/test split were manually annotated (by two annotators) to ensure high-quality evaluation. Additionally, 100 random documents from the test set were annotated twice to calculate the relevant Cohen's $\kappa$ coefficient (we achieved excellent quality $\kappa=0.9$).

\begin{table*}[htbp]
\setlength{\tabcolsep}{2pt}
\centering
\scriptsize{
\begin{tabular}{p{2,0cm}llllll}
\toprule
 Entities & General & Total & Unique & (*)Avg. entity & (*)Avg. token & Example gold value \\
 & entity type & count & values & count/doc & count/entity & \\
 \midrule
 \multicolumn{7}{c}{\emph{NDA} dataset (540 documents)} \\
 \midrule
 party & ORG/PER & 1,035 & 912 & 19.74 & 1.62 & Ajinomoto Althea Inc.  \\
 jurisdiction & LOCATION & 531 & 37 & 1.05 & 1.21 & New York  \\
 effective\_date & DATE & 400 & 370 & 1.95 & 3.10 & 2005-07-03 \\
 term & DURATION & 194 & 22 & 1.03 & 2.77 & P12M \\
 \midrule
 \multicolumn{7}{c}{\emph{Charity} dataset (2 788 documents)}\\
 \midrule
 post\_town & ADDRESS & 2,692 & 501 & 1.12 & 1.06 & BURY \\
 postcode & ADDRESS & 2,717 & 1,511 & 1.12 & 1.99 & BL9 ONP \\
 street\_line & ADDRESS & 2,414 & 1,353 & 1.12 & 2.52 & 42-47 MINORIES \\
 charity\_name & ORG & 2,778 & 1,600 & 13.80 & 3.67 & Mad Theatre Company \\
 charity\_number & NUMBER & 2,763 & 1,514 & 2.47 & 1.00 & 1143209 \\
 report\_date & DATE & 2,776 & 129 & 10.58 & 2.96 & 2016-09-30 \\
 income & AMOUNT & 2,741 & 2,726 & 1.95 & 1.01 & 109370.00 \\
 spending & AMOUNT & 2,731 & 2,712 & 2.03 & 1.01 & 90174.00 \\
 \bottomrule
\end{tabular}
}
\caption{Summary of the entities in the NDA and Charity datasets. (*) Based on manual annotation of text spans.\label{tab:entities_stats}}
\end{table*}

\subsection{Statistics and Analysis}
\label{sec:statistics_analysis}

The detailed statistics of the Kleister datasets are presented in Table~\ref{tab:dataset_summary} and Table~\ref{tab:entities_stats}. Our datasets covered a broad range of general types of entities; the \entity{party} entity is special since it could be one of the following types: \entity{ORGANIZATION} or \entity{PERSON}. Additionally, some documents may not contain all entities mentioned in the text, for instance in Kleister NDA the \entity{term} entity appears in 36~\% of documents. Likewise, some entities may have more than one gold value; for instance in Kleister NDA the \entity{party} entity could have up to 7 gold values for a single document. \entity{Report\_date}, \entity{jurisdiction} and \entity{term} have the lowest number of unique values. This suggests that these entities should be simpler than others to extract.

\subsubsection{Manual Annotation of Text Spans.}\label{secsec:annotation_text_spans} To give more detailed statistics we decided to annotate small numbers of documents on text span level. Four annotators annotated 60/55 documents for, respectively, the Kleister Charity and Kleister NDA. In Table~\ref{tab:entities_stats}, we observe that 5 out of 12 entities appear once in a single document. There are also three entities with more than ten counts on average (\entity{party}, \entity{charity\_number} and \entity{report\_date}). Annotation on the text-span level could prove critical to checking the quality of the training dataset for methods based on a Named Entity Recognition model, something which an \emph{autotagging} mechanism produces (see section~\ref{sec:pipeline}).

\begin{figure*}[ht!]
\begin{center}
\includegraphics[width=1\textwidth]{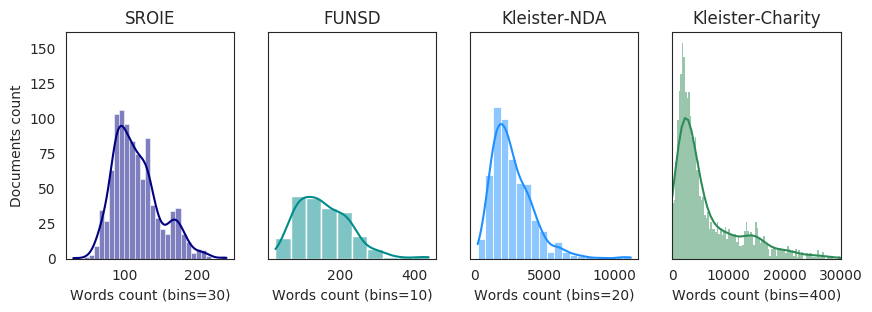}
\caption{Distribution of document lengths for Kleister datasets compared to other similar datasets (note that the x-axes ranges are different).\label{fig:histogram-doc-len}}
\end{center}
\end{figure*}

\subsubsection{Comparison with Existing Resources.} In Table~\ref{tab:dataset_summary}, we gathered the most important information about open datasets (which are the most popular ones in the domain) and the Kleister datasets. In particular, we outlined the difference based on the following properties:
\begin{itemize}
    \item \textbf{Input/Output on token level}: it is known which tokens an entity is made up from in the documents. Otherwise, one should: a) create a method for preparing a training dataset for sequence labeling models (subsequently in the publication, we use the term \textit{autotagging} for this sub-task); b) infer or create a canonical form of the final output in order to deal with differences between the target entities provided in the annotations and their variants occurring in the documents (e.g. for \entity{jurisdiction} we must transform a text-level span \textit{NY} into a document-level gold value \textbf{New York}).
    \item \textbf{Long Document}: Figure~\ref{fig:histogram-doc-len} presents differences in document lengths (calculated as a number of OCRed words) in the Kleister datasets compared to other similar datasets. Since entities could appear in documents multiple times in different contexts, we must be able to understand long documents as a whole. This leads, of course, to different architectural decisions~\cite{beltagy2020longformer,Dai_2019}. For example, the \entity{term} entity in the Kleister NDA dataset tells us about the contract duration. This information is generally found in the \textit{Term} chapter, in the middle part of a document. However, sometimes we could also find a \entity{term} entity at the end of the document, the task is to find out which of the values is correct.
    \item \textbf{Complex Layout}: this requires proper understanding of the complex layout (e.g. interpreting tables and forms as 2D structures), see Fig~\ref{fig:datasets_problems}.
    \item \textbf{OCR}: processing of scanned documents in such a way as to deal with possible OCR errors caused by handwriting, pages turned upside down or more general poor scan quality.
\end{itemize}

\begin{figure*}[ht!]
\begin{center}
\includegraphics[width=1\textwidth]{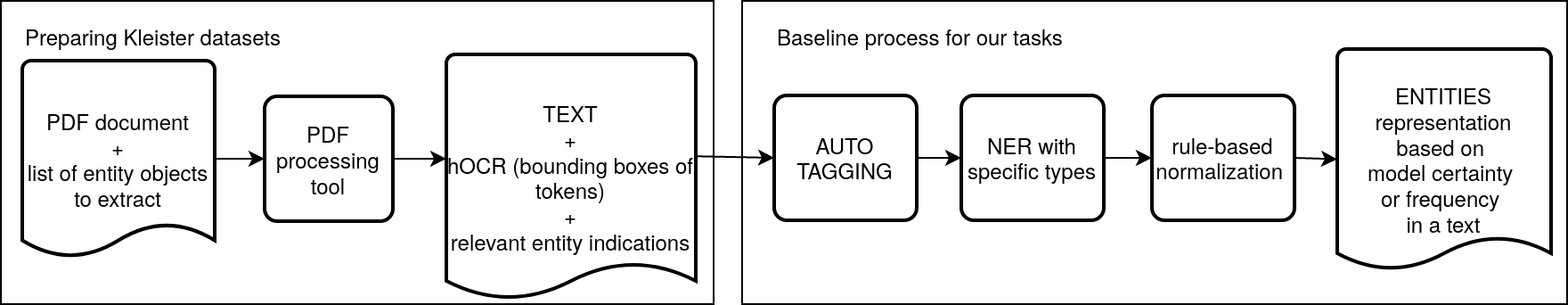}
\caption{Our preparation process for Kleister datasets and training baselines. Initially, we gathered PDF documents and expected entities' values. Then, based on textual and layout features, we prepared our pipeline solutions. The pipeline process is illustrated in the second frame and consists of the following stages: autotagging; standard NER; text normalization; and final selection of the values of entities.\label{fig:data-baselines}}
\end{center}
\end{figure*}

\section{Experiments}
\label{sec:baselines}

Kleister datasets for the Key Information Extraction task are challenging and hardly any solutions in the current NLP world can solve them. In this experiment, we aim to produce strong baselines with the \textit{Pipeline} approach (Section~\ref{sec:pipeline}) to solve extraction problems. This method's core idea is to select specific parts of the text in a document that denote the objects we search for. The whole process is a chain of techniques with, crucially, a named entity recognition model: once indicated in a document (multiple occurrences are possible), entities are normalized, then all results are aggregated into the one value specified for a given entity type.

\subsection{Document Processing Pipeline}
\label{sec:pipeline}

Figure~\ref{fig:data-baselines} presents the whole process, and all the stages are described below (a similar methodology was proposed \bycite{Palm_2017}).

\textbf{Autotagging} Since we have only document-level annotation in the Kleister datasets, we need to generate a training set for an~NER model which takes text span annotation as the input. This stage involves extracting all the fragments that refer to the same or to different entities by using sets of regular expressions combined with a gold-standard value for each general entity type, e.g. date, organization and amount. In particular, when we try to detect a~{\entity{report\_date}} entity, we must handle different date formats: `November 29, 2019’, `11/29/19’ or \noqa{latex-8}`11-29-2019’. This step is performed only for the purpose of training (to get data for training a sequence labeler; it is not applied during the prediction). The quality of this step varies across entity types (see details in Table~\ref{tab:baselines}).

\textbf{Named Entity Recognition.} We trained a NER model on the autotagged dataset using one of the state-of-the-art (1) architectures working on plain text such as Flair~\cite{akbik-etal-2018-contextual}, BERT-base~\cite{DBLP-journals-corr-abs-1810-04805}, RoBERTa-base~\cite{Liu2019RoBERTaAR}, or (2) models employing layout features (LayoutLM-base~\cite{Xu_2020} and LAMBERT~\cite{garncarek2020lambert}). Then, at the evaluation stage, we use the NER model to detect all entity occurrences in the text.

\textbf{Normalization.} At this stage objects are normalized to the canonical form, which we have defined in the Kleister datasets. We use almost the same regular expression as during autotagging. For instance, all detected {\entity{report\_date}} occurrences are normalized. So `November 29, 2019’, `11/29/19’ and \noqa{latex-8}`11-29-2019’ are rendered in our standard `2019-11-29’ form (ISO~8601).

\textbf{Aggregation.} The NER model might return more than one text span for a given entity, sometimes these are multiple occurrences of the same correct information. Sometimes these represent errors of the NER model. In any case, we need to produce a single output from multiple candidates detected by the NER model. We take a simple approach: all candidates are grouped by the extracted entities' normalized forms and for each group we sum up the scores and finally we return the values with the largest sums.

\subsection{Experimental Setup}

Due to the Kleister document's length, most currently available models limit input size and so are unable to process the documents in a single pass. Therefore, each document was split into 300-word chunks (for Flair) or 510 BPE tokens (for BERT/RoBERTa/LayoutLM/LAMBERT) with overlapping parts.  The results from overlapping parts were aggregated by averaging all the scores obtained for each token in the overlap.

For the Flair-based pipeline, we used implementation from the Flair library~\cite{akbik-etal-2018-contextual} in version 0.6.1 with the following parameters: $\mathit{learning\, rate}=0.1$, $\mathit{batch\, size}=32$, $\mathit{hidden\, size}=256$, $\mathit{epoch}=30/15$ (resp. NDA and Charity), $\mathit{patience}=3$, $\mathit{anneal\, factor}=0.5$, and with a CRF layer on top. For pipeline based on BERT/RoBERTa/LayoutLM, we used the implementation from \noqa{spell-ers}\textit{transform\-ers}~\cite{pytorch-transformers} library in version 3.1.0 with the following parameters: $\mathit{learning\, rate}=2\mathrm{e}{-5}$, $\mathit{batch\, size}=8$, $\mathit{epoch}=20$, $\mathit{patience}=2$. For pipeline based on LAMBERT model we used implementation shared by authors of the publication \cite{garncarek2020lambert} and the same parameters as for the BERT/RoBERTa/LayoutLM models. All experiments were performed with the same settings.

Moreover, in our experiments, we tried different PDF processing tools for text extraction from PDF documents to check the importance of text quality for the final pipeline score:
\begin{itemize}
    \item \textbf{Microsoft Azure Computer Vision API (Azure CV)}\footnote{\url{https://docs.microsoft.com/en-us/azure/cognitive-services/computer-vision/concept-recognizing-text}} -- commercial OCR engine, version 3.0.0;
    \item \textbf{pdf2djvu/djvu2hocr}\footnote{\url{http://jwilk.net/software/pdf2djvu}, \url{https://github.com/jwilk/ocrodjvu}}-- a free tool for object and text extraction from born-digital PDF files (this is not an OCR engine, hence it could be applied only to Kleister NDA), version 0.9.8;
    \item \textbf{Tesseract}\cite{tesseract} -- this is the most popular free OCR engine currently available, we used version \noqa{spell-rc1}\noqa{spell-gb36c}4.1.1-rc1-7-gb36c.\footnote{run with \code{-{}-oem 2 -l eng -{}-dpi 300} flags (meaning both new and old OCR engines were used simultaneously, with language and pixel density set to English and 300dpi respectively)};
    \item \textbf{Amazon Textract}\footnote{\url{https://aws.amazon.com/textract/} (API in version from March 1, 2020 was used)} -- commercial OCR engine.
\end{itemize}

\begin{table*}[ht!]
\centering
\scriptsize{
\begin{tabular}{p{2.0cm}p{1.0cm}p{1.0cm}p{1.6cm}p{1.6cm}p{1.1cm}p{1.0cm}p{1.0cm}}
\toprule
\multicolumn{8}{c}{Kleister NDA dataset (pdf2djvu)} \\
\midrule
Entity name & \multicolumn{1}{l}{\textbf{Flair}} & \multicolumn{1}{l}{\textbf{BERT}} & \multicolumn{1}{l}{\textbf{RoBERTa}}&
\multicolumn{1}{l}{\textbf{LayoutLM}}&
\multicolumn{1}{l}{\textbf{LAMBERT}}&
\multicolumn{1}{l}{Autotagger}&
\multicolumn{1}{l}{Human}\\
\midrule
effective\_date & 79.37 & 80.20 & 81.50 & 80.50 & \textbf{85.27} & 79.00 & 100\% \\
party & 70.13 & 71.60 & \textbf{80.83} & 76.60 & 78.70 & 33.15 & 98\% \\
jurisdiction & 93.87 & 95.00 & 92.87 & 94.23 & \textbf{96.50} & 54.10 & 100\% \\
term & \textbf{60.33} & 45.73 & 52.27 & 47.63 & 55.03 & 74.10 & 95\% \\
\midrule
ALL & 77.83 & 78.20 & 81.00 & 78.47 & \textbf{81.77} & 60.09 & 97.86\% \\
\toprule
\multicolumn{8}{c}{Kleister Charity dataset (Azure CV)} \\
\midrule
post\_town & 83.07 & 77.03 & 77.70 & 76.57 &\textbf{83.70} & 66.04 & 98\% \\
postcode & 89.57 & 87.10 & 88.40 & 88.53 & \textbf{90.37} & 87.60 & 100\% \\
street\_line & 69.10 & 62.23 & 72.03 & 70.92 & \textbf{74.30} & 75.02 & 96\% \\
charity\_name & 72.97 & 75.93 & 78.03 & \textbf{79.63} & 77.83 & 67.00 & 99\% \\
charity\_number & 96.60 & \textbf{96.67} & 95.37 & 96.13 & 95.80 & 98.60 & 98\% \\
income & 70.67 & 67.30 & 69.73 & 70.40 & \textbf{74.70} & 69.00 & 97\% \\
report\_date & 95.93 & 96.60 & 96.77 & 96.40 & \textbf{96.80} & 89.00 & 100\% \\
spending & 68.13 & 64.43 & 68.60 & 68.57 & \textbf{74.20} & 73.00 & 92\% \\
\midrule
ALL & 81.17 & 78.33 & 81.50 & 81.53 & \textbf{83.57} & 78.16 & 97.45\% \\
\bottomrule
\end{tabular}
}
\vspace{1mm}
\caption{The detailed results (average $F_1$-scores over 3 runs) of our baselines for Kleister challenges (test sets) for the best PDF processing tool. Autotagger $F_1$-scores were calculated based on results from our regexp mechanism and manual annotation on the text span level (see section~\ref{secsec:annotation_text_spans}). Human performance is a percentage of annotators agreements for 100 random documents. We used the Base version of the BERT, RoBERTa, LayoutLM and LAMBERT models.\label{tab:baselines}}
\end{table*}

\section{Results}\label{sec:baseline_results}

Table~\ref{tab:baselines} shows the results for the two Kleister datasets obtained with the Pipeline method for all tested models. The weakest model from our baselines is, in general, BERT, with a slight advantage in Kleister NDA over the Flair model and a large performance drop on Kleister Charity in comparison to others. The best model is LAMBERT, which improved the overall $F_1$-score with 0.77 and 2.04 for, respectively, NDA and Charity. It is worth noting that for born-digital documents in Kleister NDA this difference is not substantial. This is due to the fact that only for \entity{effective\_date} entity does the LAMBERT model have a clear advantage (about 4 points gain of $F_1$-score) over other baseline models. For Kleister Charity LAMBERT achieves the biggest improvement over sequential models on \entity{income} (+4.03) and \entity{spending} (+5.60) which appears mostly in table-like structures.

The most challenging problems for all models are entities (\entity{effective\_date}, \entity{party}, \entity{term}, \entity{post\_town}, \entity{postcode}, \entity{street\_line}, \entity{charity\_name}, \entity{income}, \entity{spending}) related to the properties described in Section \ref{sec:statistics_analysis}.

\begin{figure}
\centering
\begin{minipage}{.47\textwidth}
  \centering
  \includegraphics[width=1.05\linewidth]{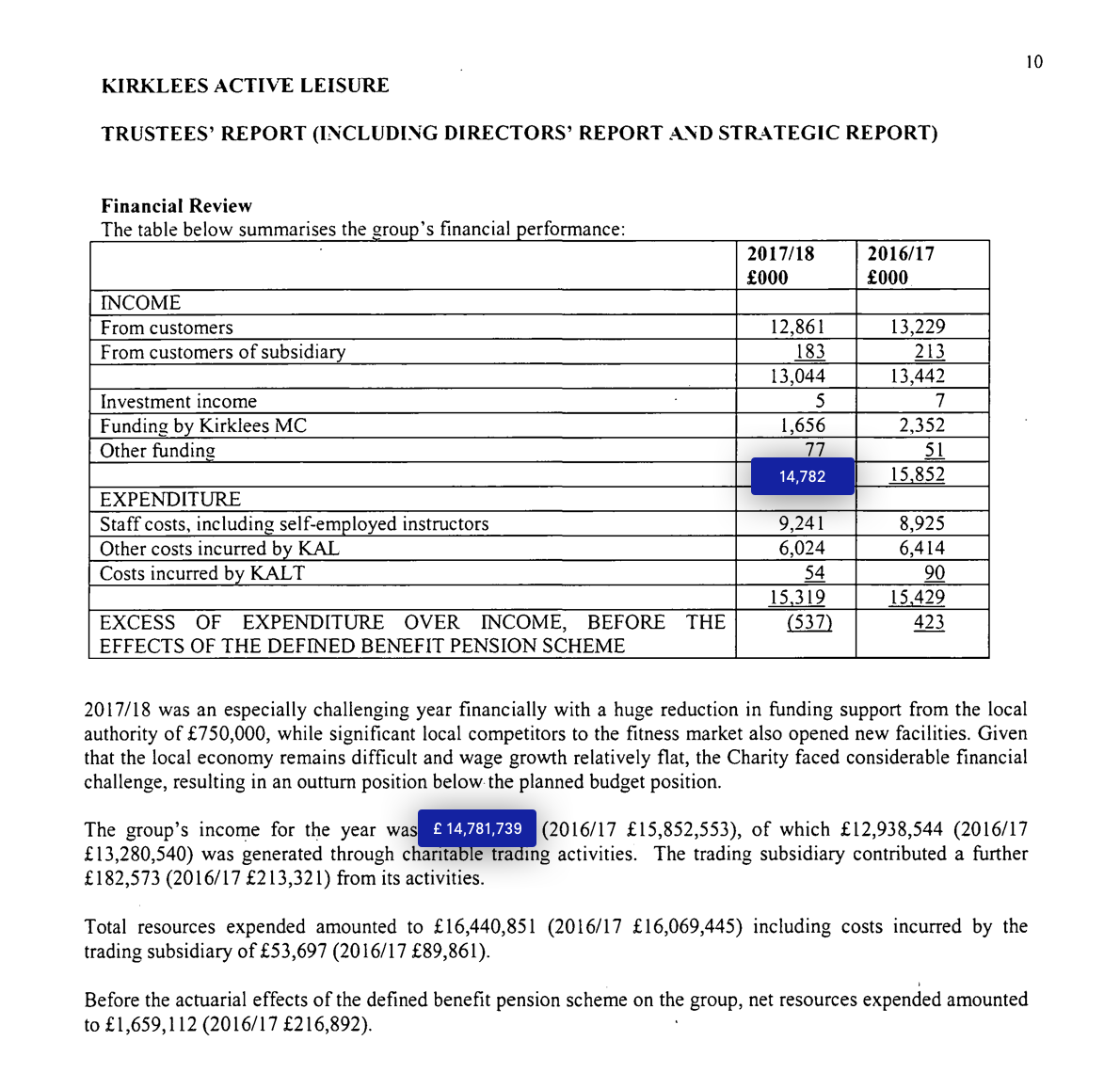}
  \caption{Normalization issues for an \entity{income} entity (amount in the table should be multiplied by 1000).\label{fig:charity_normalization}}
\end{minipage}%
\begin{minipage}{.04\textwidth}
\ %
\end{minipage}%
\begin{minipage}{.47\textwidth}
  \centering
  \includegraphics[width=.99\linewidth]{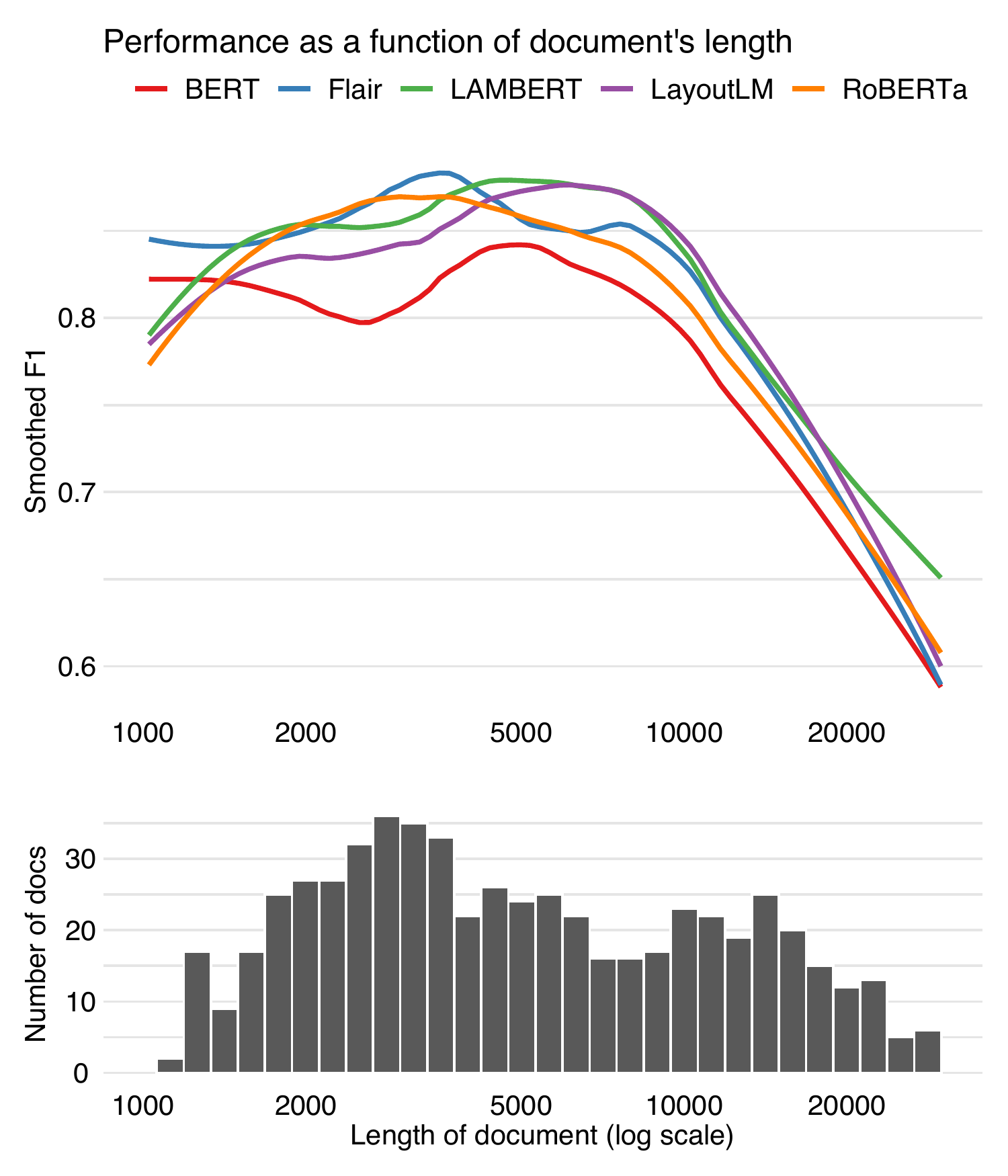}
  \caption{Relationship between $F_1$-scores and document length in the Kleister Charity test set for the Azure CV OCR.}
\label{fig:doc_word_count2f_score}
\end{minipage}
\end{figure}

\textbf{Input/Output on Token Level (Autotagging).} As we can observe in Table~\ref{tab:baselines}, our autotagging mechanism with information about entity achieves, on the text span level, a performance inferior to almost all our models on the document level. It shows that, despite the fact that the autotagging mechanism is prone to errors, we could train a good quality NER model. Our analysis shows that there are some specific issues related to a regular-expression-based mechanism, e.g. \entity{party} in the Kleister NDA dataset has the lowest score because organization names often occur in the text as an acronym or as a shortened form; for instance for \entity{party} entity text \textit{Emerson Electric Co.} means the same as \textit{Emerson}. This is not easy to capture with a general regexp rule.

\textbf{Input/Output on token level (normalization).} We found that we could not achieve competitive results by using models based only on sequence labeling. For example, for the entities \entity{income} and \entity{spending} in the Kleister Charity dataset, we manually checked that in about 5~\% of examples we need to also infer the right scale (thousand, million, etc.) for each monetary value based on the document context (see Figure~\ref{fig:charity_normalization}).

\textbf{Long Documents.} It turns out that, for all models, worse results are observed for longer documents, see Figure~\ref{fig:doc_word_count2f_score}.

\textbf{Complex Layout.} The LAMBERT model has proved the best one, which proved the importance of using models employing not only textual (1D) but also layout (2D) features (see Table~\ref{tab:baselines}). Additionally, we also observe that the entities appearing in the sequential contexts achieve higher $F_1$-scores (\entity{charity\_number} and \entity{report\_date} entities in the Kleister Charity dataset).

\textbf{OCR.} We present the importance of using a PDF processing tool of good quality (see Table~\ref{tab:ocr_baselines}). With such a tool, we could gain several points in the $F_1$-score. There are two main conclusions: 1) Commercial OCR engines (Azure CV and Textract) are significantly better than Tesseract for scanned documents (Kleister Charity dataset). This is especially for true for 1D models not trained on Tesseract output (Flair, BERT, RoBERTa); 2) If we have the means to detect born-digital PDF documents, we should process them with a dedicated PDF tool (such as pdf2djvu) instead of using an OCR engine.

\begin{table*}[ht!]
\centering
\small{
\begin{tabular}{p{1.9cm}p{1.9cm}p{1.9cm}p{1.9cm}p{1.9cm}p{1.9cm}}
\toprule
\multicolumn{6}{c}{Kleister NDA dataset (born-digital PDF files)}\\
\midrule
PDF tool & \multicolumn{1}{l}{\textbf{Flair}} & \multicolumn{1}{l}{\textbf{BERT}} & \multicolumn{1}{l}{\textbf{RoBERTa}}& \multicolumn{1}{l}{\textbf{LayoutLM}}&
\multicolumn{1}{l}{\textbf{LAMBERT}}\\
\midrule
Azure CV & 78.03\textsubscript{$\pm$0.12} & 77.67\textsubscript{$\pm$0.18} & 79.33\textsubscript{$\pm$0.68} & 77.43\textsubscript{$\pm$0.29} & 80.57\textsubscript{$\pm$0.25}\\
pdf2djvu & 77.83\textsubscript{$\pm$0.26} & 78.20\textsubscript{$\pm$0.17} & 81.00\textsubscript{$\pm$0.05} & 78.47\textsubscript{$\pm$0.76} & \textbf{81.77\textsubscript{$\pm$0.09}}\\
Tesseract & 76.57\textsubscript{$\pm$0.49} & 76.60\textsubscript{$\pm$0.30} & 77.81\textsubscript{$\pm$0.97} & 77.70\textsubscript{$\pm$0.48} & 81.03\textsubscript{$\pm$0.23}\\
Textract & 77.37 \textsubscript{$\pm$0.08} & 74.83\textsubscript{$\pm$0.45} & 79.49\textsubscript{$\pm$0.32} & 77.40\textsubscript{$\pm$0.40} & 77.37\textsubscript{$\pm$0.08}\\
\toprule\multicolumn{6}{c}{Kleister Charity dataset (mixture of born-digital and scanned PDF files) (*)} \\
\midrule
Azure CV & 81.17\textsubscript{$\pm$0.12} & 78.33\textsubscript{$\pm$0.08} & 81.50\textsubscript{$\pm$0.23} & 81.53\textsubscript{$\pm$0.23} &
\textbf{83.57\textsubscript{$\pm$0.29}} \\
Tesseract & 72.87\textsubscript{$\pm$0.81} & 71.37\textsubscript{$\pm$1.25} & 76.23\textsubscript{$\pm$0.15} & 77.53\textsubscript{$\pm$0.20} & 81.50\textsubscript{$\pm$0.07} \\
Textract & 78.03\textsubscript{$\pm$0.12} & 73.30\textsubscript{$\pm$0.43} & 80.08\textsubscript{$\pm$0.15} & 80.23\textsubscript{$\pm$0.41} &
82.97\textsubscript{$\pm$0.21} \\
\bottomrule
\end{tabular}
}
\vspace{1mm}
\caption{$F_1$-scores for different PDF processing tools and models checked on Kleister challenges test sets over 3 runs with standard deviation. (*) pdf2djvu does not work on scans. We used the Base version of the BERT, RoBERTa, LayoutLM and LAMBERT models.\label{tab:ocr_baselines}}
\end{table*}

\section{Conclusions}
\label{sec:conclusions}

In this paper, we introduced two new datasets Kleister NDA and
Kleister Charity for Key Information Extraction tasks. We set out in
detail the process necessary for the preparation of these datasets. Our intention was
to show that Kleister datasets will help the NLP community to investigate the
effects of document lengths, complex layouts, and OCR quality
problems on KIE performance.

We prepared baseline solutions based on text and layout data generated by different PDF processing tools from the datasets. The best model from our baselines achieves 81.77/83.57 $F_1$-score for, respectively, the Kleister NDA and Charity, which is much lower in comparison to datasets in a similar domain (e.g. 98.17~\cite{garncarek2020lambert} for SROIE). This benchmark shows the weakness of the currently available state-of-the-art models for the Key Information Extraction task.

\bibliography{ms}

\begin{thebibliography}{10}
\providecommand{\url}[1]{\texttt{#1}}
\providecommand{\urlprefix}{URL }
\providecommand{\doi}[1]{https://doi.org/#1}

\bibitem{akbik-etal-2018-contextual}
Akbik, A., Blythe, D., Vollgraf, R.: Contextual string embeddings for sequence
  labeling. In: Proceedings of the 27th International Conference on
  Computational Linguistics. pp. 1638--1649. Association for Computational
  Linguistics, Santa Fe, New Mexico, USA (Aug 2018),
  \url{https://www.aclweb.org/anthology/C18-1139}

\bibitem{beltagy2020longformer}
Beltagy, I., Peters, M.E., Cohan, A.: Longformer: The long-document
  transformer. ArXiv  \textbf{abs/2004.05150} (2020)

\bibitem{DBLP:conf/emnlp/BorchmannWGKJSP20}
Borchmann, L., Wisniewski, D., Gretkowski, A., Kosmala, I., Jurkiewicz, D.,
  Szalkiewicz, L., Palka, G., Kaczmarek, K., Kaliska, A., Gralinski, F.:
  Contract discovery: Dataset and a few-shot semantic retrieval challenge with
  competitive baselines. In: Cohn, T., He, Y., Liu, Y. (eds.) Proceedings of
  the 2020 Conference on Empirical Methods in Natural Language Processing:
  Findings, {EMNLP} 2020, Online Event, 16-20 November 2020. pp. 4254--4268.
  Association for Computational Linguistics (2020)

\bibitem{Dai_2019}
Dai, Z., Yang, Z., Yang, Y., Carbonell, J., Le, Q., Salakhutdinov, R.:
  Transformer-xl: Attentive language models beyond a fixed-length context.
  Proceedings of the 57th Annual Meeting of the Association for Computational
  Linguistics  (2019), \url{https://www.aclweb.org/anthology/P19-1285}

\bibitem{DBLP-journals-corr-abs-1810-04805}
Devlin, J., Chang, M., Lee, K., Toutanova, K.: {BERT:} pre-training of deep
  bidirectional transformers for language understanding. ArXiv
  \textbf{abs/1810.04805} (2018)

\bibitem{dwojak-etal-2020-dataset}
Dwojak, T., Pietruszka, M., Borchmann, {\L}., Ch{\l}{\k{e}}dowski, J.,
  Grali{\'n}ski, F.: From dataset recycling to multi-property extraction and
  beyond. In: Proceedings of the 24th Conference on Computational Natural
  Language Learning. pp. 641--651. Association for Computational Linguistics,
  Online (Nov 2020). \doi{10.18653/v1/2020.conll-1.52},
  \url{https://www.aclweb.org/anthology/2020.conll-1.52}

\bibitem{garncarek2020lambert}
Garncarek, {\L}., Powalski, R., Stanisławek, T., Topolski, B., Halama, P.,
  Turski, M., Graliński, F.: {LAMBERT}: {Layout-Aware} ({Language}) {Modeling}
  using {BERT} for information extraction. ArXiv  \textbf{abs/2002.08087}
  (2020)

\bibitem{hewlett-etal-2016-wikireading}
Hewlett, D., Lacoste, A., Jones, L., Polosukhin, I., Fandrianto, A., Han, J.,
  Kelcey, M., Berthelot, D.: {W}iki{R}eading: A novel large-scale language
  understanding task over {W}ikipedia. In: Proceedings of the 54th Annual
  Meeting of the Association for Computational Linguistics (Volume 1: Long
  Papers). pp. 1535--1545. Association for Computational Linguistics, Berlin,
  Germany (2016)

\bibitem{holt-chisholm-2018-extracting}
Holt, X., Chisholm, A.: Extracting structured data from invoices. In:
  Proceedings of the Australasian Language Technology Association Workshop
  2018. pp. 53--59. Dunedin, New Zealand (Dec 2018),
  \url{https://www.aclweb.org/anthology/U18-1006}

\bibitem{pytorch-transformers}
{Hugging Face}: Transformers. \url{https://github.com/huggingface/transformers}
  (2020)

\bibitem{jaume2019}
{Jaume}, G., {Kemal Ekenel}, H., {Thiran}, J.: {FUNSD}: A dataset for form
  understanding in noisy scanned documents. In: 2019 International Conference
  on Document Analysis and Recognition Workshops (ICDARW). vol.~2, pp.~1--6
  (2019)

\bibitem{DBLP-journals-corr-abs-1809-08799}
Katti, A.R., Reisswig, C., Guder, C., Brarda, S., Bickel, S., H{\"{o}}hne, J.,
  Faddoul, J.B.: {Chargrid: Towards Understanding 2D Documents}. ArXiv
  \textbf{abs/1809.08799} (2018)

\bibitem{Liu_2019}
Liu, X., Gao, F., Zhang, Q., Zhao, H.: Graph convolution for multimodal
  information extraction from visually rich documents. Proceedings of the 2019
  Conference of the North American Chapter of the Association for Computational
  Linguistics  (2019), \url{http://dx.doi.org/10.18653/v1/N19-2005}

\bibitem{Liu2019RoBERTaAR}
Liu, Y., Ott, M., Goyal, N., Du, J., Joshi, M., Chen, D., Levy, O., Lewis, M.,
  Zettlemoyer, L., Stoyanov, V.: {RoBERTa: A Robustly Optimized BERT
  Pretraining Approach}. ArXiv  \textbf{abs/1907.11692} (2019)

\bibitem{mathew2021docvqa}
Mathew, M., Karatzas, D., Jawahar, C.V.: {DocVQA: A Dataset for VQA on Document
  Images}. ArXiv  \textbf{abs/2007.00398} (2021)

\bibitem{Palm_2019}
Palm, R.B., Laws, F., Winther, O.: Attend, copy, parse end-to-end information
  extraction from documents. International Conference on Document Analysis and
  Recognition (ICDAR)  (2019)

\bibitem{Palm_2017}
Palm, R.B., Winther, O., Laws, F.: Cloudscan - a configuration-free invoice
  analysis system using recurrent neural networks. 14th IAPR International
  Conference on Document Analysis and Recognition (ICDAR)  (2017).
  \doi{10.1109/icdar.2017.74}

\bibitem{park2019cord}
Park, S., Shin, S., Lee, B., Lee, J., Surh, J., Seo, M., Lee, H.: {CORD: A
  Consolidated Receipt Dataset for Post-OCR Parsing}. Document Intelligence
  Workshop at Neural Information Processing Systems  (2019)

\bibitem{tesseract}
Smith, R.: {Tesseract Open Source OCR Engine} (2020),
  \url{https://github.com/tesseract-ocr/tesseract}

\bibitem{W03-0419}
Tjong Kim~Sang, E.F., De~Meulder, F.: {Introduction to the CoNLL-2003 Shared
  Task: Language-Independent Named Entity Recognition}. In: Proceedings of the
  Seventh Conference of the North American Chapter of the Association for
  Computational Linguistics (2003)

\bibitem{Wellmann_2020}
Wellmann, C., Stierle, M., Dunzer, S., Matzner, M.: A framework to evaluate the
  viability of robotic process automation for business process activities.
  Business Process Management: Blockchain and Robotic Process Automation Forum
  (2020)

\bibitem{DBLP-conf-fedcsis-WroblewskaSPG18}
Wróblewska, A., Stanisławek, T., Prus{-}Zajączkowski, B., Garncarek, {\L}.:
  Robotic process automation of unstructured data with machine learning. In:
  Position Papers of the 2018 Federated Conference on Computer Science and
  Information Systems, FedCSIS 2018, Pozna{\'{n}}, Poland, September 9-12,
  2018. pp. 9--16 (2018). \doi{10.15439/2018F373}

\bibitem{Xu_2020}
Xu, Y., Li, M., Cui, L., Huang, S., Wei, F., Zhou, M.: {LayoutLM}: Pre-training
  of text and layout for document image understanding. Proceedings of the 26th
  ACM SIGKDD International Conference on Knowledge Discovery \& Data Mining
  (2020). \doi{10.1145/3394486.3403172}

\end{thebibliography}
\bibliographystyle{splncs04}

\ifwithappendix
\input{appendix}
\fi

\end{document}